\documentclass[letterpaper, 10 pt, conference]{ieeeconf}  %

\IEEEoverridecommandlockouts                           

\usepackage{times}
\usepackage{microtype}
\usepackage{epsfig}
\usepackage[table,xcdraw]{xcolor}
\usepackage{caption}
\usepackage{float}
\usepackage{placeins}
\usepackage{color, colortbl}
\usepackage{stfloats}
\usepackage{tabularx}
\usepackage{xstring}
\usepackage{multirow}
\usepackage{booktabs}
\usepackage{xspace}
\usepackage{url}
\usepackage{subcaption}
\usepackage{xcolor}
\usepackage[hang,flushmargin]{footmisc}
\usepackage{algorithm}
\usepackage[noend]{algpseudocode}
\usepackage{amsfonts}
\usepackage{amsmath}
\usepackage{bbm}
\usepackage{hyperref}
\usepackage{amssymb}
\usepackage{dsfont}

\makeatletter
\DeclareRobustCommand\onedot{\futurelet\@let@token\@onedot}
\def\@onedot{\ifx\@let@token.\else.\null\fi\xspace}

\def\etal{\emph{et al}\onedot}
\makeatother

\newcommand{\csz}[1]{{\textcolor{black}{#1}}}

\newcommand{\R}[1]{{%
    \textbf{%
        \ifstrequal{#1}{1}{\textcolor{red}{R#1}}{%
        \ifstrequal{#1}{2}{\textcolor{blue}{R#1}}{%
        \ifstrequal{#1}{3}{\textcolor{magenta}{R#1}}{%
        \ifstrequal{#1}{4}{\textcolor{teal}{R#1}}{%
                           \textcolor{cyan}{R#1}%
        }}}}%
    }%
}}

\title{\LARGE \bf
Object Goal Navigation with Recursive Implicit Maps
}

\author{Shizhe Chen$^1$, Thomas Chabal$^1$, Ivan Laptev$^1$ and Cordelia Schmid$^1$%
\thanks{$^1$Inria, \'Ecole normale sup\'erieure, CNRS, PSL Research University, 75005, Paris, France. {\tt\small \{firstname.lastname\}@inria.fr}}\\
}

\begin{document}

\bstctlcite{IEEEexample:BSTcontrol}

\maketitle
\thispagestyle{empty}
\pagestyle{empty}

\begin{abstract}
Object goal navigation aims to navigate an agent to locations of a given object category in unseen environments.
Classical methods explicitly build maps of environments and require extensive engineering while lacking semantic information for object-oriented exploration.
On the other hand, end-to-end learning methods alleviate manual map design and predict actions using implicit representations.
Such methods, however, lack an explicit notion of geometry and may have limited ability to encode navigation history.
In this work, we propose an implicit spatial map for object goal navigation.
Our implicit map is recursively updated with new observations at each step using a transformer.
To encourage spatial reasoning, we introduce auxiliary tasks and train our model to reconstruct explicit maps as well as to predict visual features, semantic labels and actions.
Our method significantly outperforms the state of the art on the challenging MP3D dataset and generalizes well to the HM3D dataset.
We successfully deploy our model on a real robot and achieve encouraging object goal navigation results in real scenes using only a few real-world demonstrations.
Code, trained models and videos are available at \url{https://www.di.ens.fr/willow/research/onav_rim/}.
\end{abstract}
\section{Introduction}
\label{sec:intro}

Searching for objects in the real world is an essential ability for embodied agents and serves as a precursor to many object manipulation tasks.
As a consequence, many recent works have focused on the object goal navigation (ObjectNav)~\cite{batra2020objectnav,chaplot2020object,maksymets2021thda,Ye_2021_ICCV,ramrakhya2022habitatweb,ramakrishnan2022poni}.
This task requires an agent to navigate in unknown environments and to reach locations next to an instance of the goal object category. 

The ObjectNav task is more challenging compared to the classic visual navigation to a given location~\cite{wijmans2019dd,desouza2002survey}.
Indeed, in addition to finding a collision-free path, ObjectNav also requires to reason about the semantics in the visual world and to develop efficient object-oriented exploration strategies for previously unseen environments.
Recently, much progress has been made to improve egocentric visual perception~\cite{khandelwal2022simple,chaplot2020semantic,chaplot2021seal,yadav2022offline} thanks to the advanced visual models~\cite{he2016deep,he2017mask,dosovitskiy2020image} and large-scale self-supervised pretraining~\cite{radford2021learning}.
Yet, efficiently exploring the world remains difficult as it requires a long-term scene memory to enable backtracking as well as commonsense knowledge of scene arrangements. 
For example, a cushion is more likely to be found in the living room than in the kitchen, hence, if searching for a cushion, the exploration should focus more on the former.
Classical methods \cite{yamauchi1997frontier} using occupancy maps can be complemented with semantic recognition for ObjectNav. However, the lack of learning prevents such methods from acquiring the aforementioned commonsense knowledge.
To address this limitation, learning-based approaches~\cite{chaplot2020object,ramrakhya2022habitatweb} have been proposed to determine object-oriented navigation policies.

\begin{figure}
    \centering \includegraphics[width=0.9\linewidth]{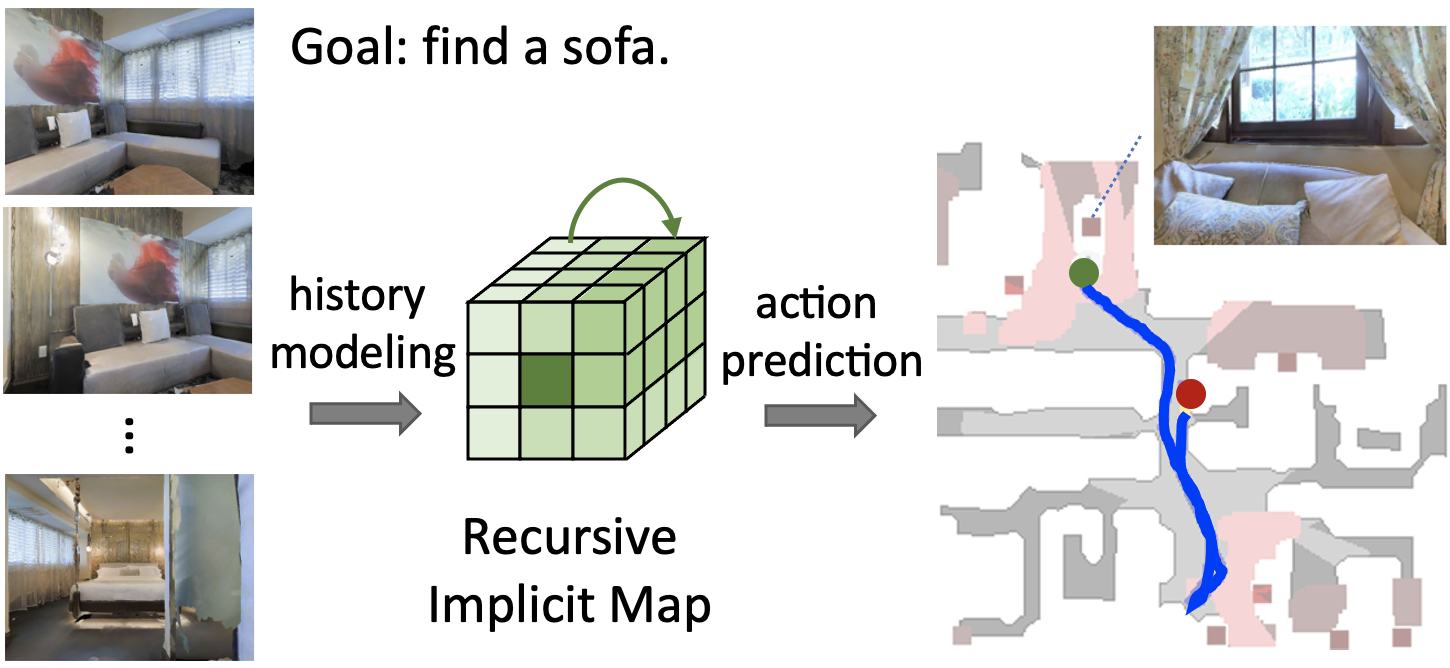}
    \caption{We propose a recursive implicit map (RIM) for object goal navigation. RIM implicitly learns the structure and texture of the environment from the navigation history, and is recursively updated to predict actions.
    The red and green circles indicate the start and stop locations respectively, while the blue curve is the trajectory of the agent. 
    Note that the agent only observes the images of the environment and does not have access to the floor map. 
    } 
    \label{fig:teaser}
    \vspace{-2em}
\end{figure}

Existing learning-based approaches to ObjectNav  can be divided into two categories, namely modular methods and end-to-end learning methods.
Modular methods~\cite{chaplot2020object,ramakrishnan2022poni,zhu2022navigating} extend occupancy maps to semantic maps by adding object dimensions to each map cell. Such methods typically deploy a neural network module to predict goal locations and another module that generates low-level actions towards the goal.
Semantic maps, however, do not scale well to many objects and are sensitive to noisy sensor measurements.
End-to-end learning methods~\cite{Ye_2021_ICCV,ramrakhya2022habitatweb,khandelwal2022simple} directly map raw observations to low-level actions.
Such methods are more flexible and can easily benefit from advances in visual representation learning~\cite{yadav2022offline,radford2021learning}.
Nevertheless, existing policy models mainly adopt recurrent neural networks (RNNs)~\cite{hochreiter1997long,chung2014empirical}. Since trajectories of low-level actions are usually long with hundreds of small steps, RNNs struggle to maintain  a long-term memory of explored environments.
A few works~\cite{pashevich2021episodic,chen2021history} proposed an episodic sequence memory using transformers~\cite{vaswani2017attention}.
Sequence representations in such models, however, fail to capture geometric structure of the environment and become inefficient for a large number of navigation steps.

In this work, we focus on end-to-end learning-based methods and propose to build a recursive implicit map (RIM) to improve object-oriented exploration, see Figure~\ref{fig:teaser}.
The implicit map consists of a spatial grid initialized from scratch at the beginning of a navigation episode. It is recursively updated with new observations by a learnable module and encodes the explored environment in geometry-aware latent vectors. Specifically, we use a multi-layer transformer model as the learnable module. The transformer only relies on the geometry and visual features to align the implicit map and observation for each update, discarding any other domain priors for map construction.
In order to improve spatial reasoning and to better supervise the learning from our implicit map,
we propose auxiliary tasks besides the primary task of action prediction. The auxiliary tasks aim to predict visual representations given the agent's pose, reconstruct a 2D occupancy map, and perform semantic recognition.

We perform extensive experiments on the Matterport3D (MP3D) ObjectNav dataset with human demonstrations~\cite{ramrakhya2022habitatweb} for training.
Experimental results demonstrate that the RIM outperforms other implicit representations including recurrent state and episodic sequence as well as explicit mapping methods.
Given outputs for auxiliary tasks, we observe that the implicit map is able to capture the geometry and semantic information in explored environments.
The proposed model achieves state-of-the-art performance on the MP3D dataset and %
generalizes well to new environments of the HM3D dataset.
We further deploy our RIM model learned in simulation in the real world with a different robot and testing environments.
Our model achieves encouraging performance after fine-tuning on few real-world demonstrations. %

\section{Related Work}
\label{sec:related}

\noindent\textbf{Visual navigation.}
The history of visual navigation can be dated back to the late 1960s with the development of the robot Shakey~\cite{nillson1984shakey, kuipers2017shakey} which combined perception and planning to accomplish high-level tasks. The field gained traction in the early 1980s \cite{desouza2002survey} and, since then, a range of visual navigation tasks \cite{boninfont2008survey,anderson2018evaluation} have been proposed for navigation in 3D environments with different goal specifications.
For example, point goal navigation (PointNav)~\cite{wijmans2019dd,savva2019habitat} uses coordinates relative to the agent's start position as the target; image goal navigation (ImageNav)~\cite{chaplot2020neural,savinov2018semi,zhu2017target} requires agents to go to the place where an image is taken; 
and in vision-and-language navigation (VLN) \cite{anderson2018vision,krantz2020beyond,shridhar2020alfred}, agents must follow step-by-step instructions to reach the goal. 
Object goal navigation (ObjectNav)~\cite{batra2020objectnav} uses the object category as target. It is particularly challenging as it requires semantic recognition to detect the goal 
and needs more efficient exploration than VLN.
While current methods have nearly `solved' PointNav~\cite{wijmans2019dd,partsey2022mapping} even in the real world and perform well in other visual navigation tasks~\cite{al2022zero,majumdarzson}, they are less competitive in ObjectNav.
Hence, in this work, we focus on improving models for ObjectNav.

\noindent\textbf{Object goal navigation.}
Modeling previous navigation history is crucial for efficient exploration and navigation in the environment.
It is typically done in two ways: explicit mapping and implicit representations.
Explicit mapping based approaches build maps for planning, as for example occupancy maps~\cite{chaplotlearning} and semantic maps~\cite{chaplot2020object,ramakrishnan2022poni, zhu2022navigating}.
Different exploration strategies have been proposed based on these maps such as frontier-based search~\cite{yamauchi1997frontier}, maximizing exploration area~\cite{chaplotlearning} and object location prediction~\cite{ramakrishnan2022poni,zhu2022navigating}.
Yet, recent work~\cite{luo2022stubborn} observes that these maps are still insufficient for efficient object-oriented exploration.
The other line of work implicitly learns representations of the environment in an end-to-end manner. Many works~\cite{maksymets2021thda,Ye_2021_ICCV,ramrakhya2022habitatweb} utilize  recurrent neural networks~\cite{hochreiter1997long,chung2014empirical}, while few works~\cite{pashevich2021episodic,chen2021history} treat navigation episodes as sequences and apply transformers to leverage long-term memory. 
To train these navigation models, reinforcement learning (RL)~\cite{sutton2018reinforcement} is the most widely used approach.
However, RL often suffers from poor sample efficiency~\cite{wijmans2019dd}, requires reward engineering and is prone to overfit in seen environments~\cite{maksymets2021thda,Ye_2021_ICCV}.
More recently, Habitat-web \cite{ramrakhya2022habitatweb} designed a web interface to easily collect human demonstrations and showed that behavior cloning with human demos enables learning smarter exploration strategies than RL.
We use this human demonstration dataset for training with behavior cloning and design a new implicit representation for end-to-end learning.

\noindent\textbf{Maps for navigation.}
Explicit maps have shown to be beneficial for navigation, especially for long-term navigation tasks \cite{wani2020multion}. 
They can be represented by occupancy maps~\cite{chaplotlearning}, topological graphs~\cite{chaplot2020neural}, meshes~\cite{rosinol2021kimera}, point clouds~\cite{jatavallabhula2023conceptfusion} or sparse points~\cite{engel2018dso}.
Such representations aim to capture the geometry of the environment for obstacle avoidance and have been extended to semantic maps \cite{chaplot2020object, ramakrishnan2022poni, zhu2022navigating}.
These maps are commonly built using Simultaneous Localization and Mapping (SLAM)~\cite{thrun2005proba}, but this construction process is tedious, notably due to noise in the sensors, expensive compute, parameter tuning and high storage requirements as the maps grow. 
Implicit scene representation is one promising alternative to alleviate these limitations: Partsey \etal \cite{partsey2022mapping} demonstrate that explicit mapping may not be necessary for navigation, even in a realistic setting, while
Wijmans \etal \cite{wijmans2023emergence} show that maps can emerge in implicit representations.

\section{Method}
\label{sec:method}

\begin{figure*}[t]
    \centering
    \includegraphics[width=0.92\linewidth]{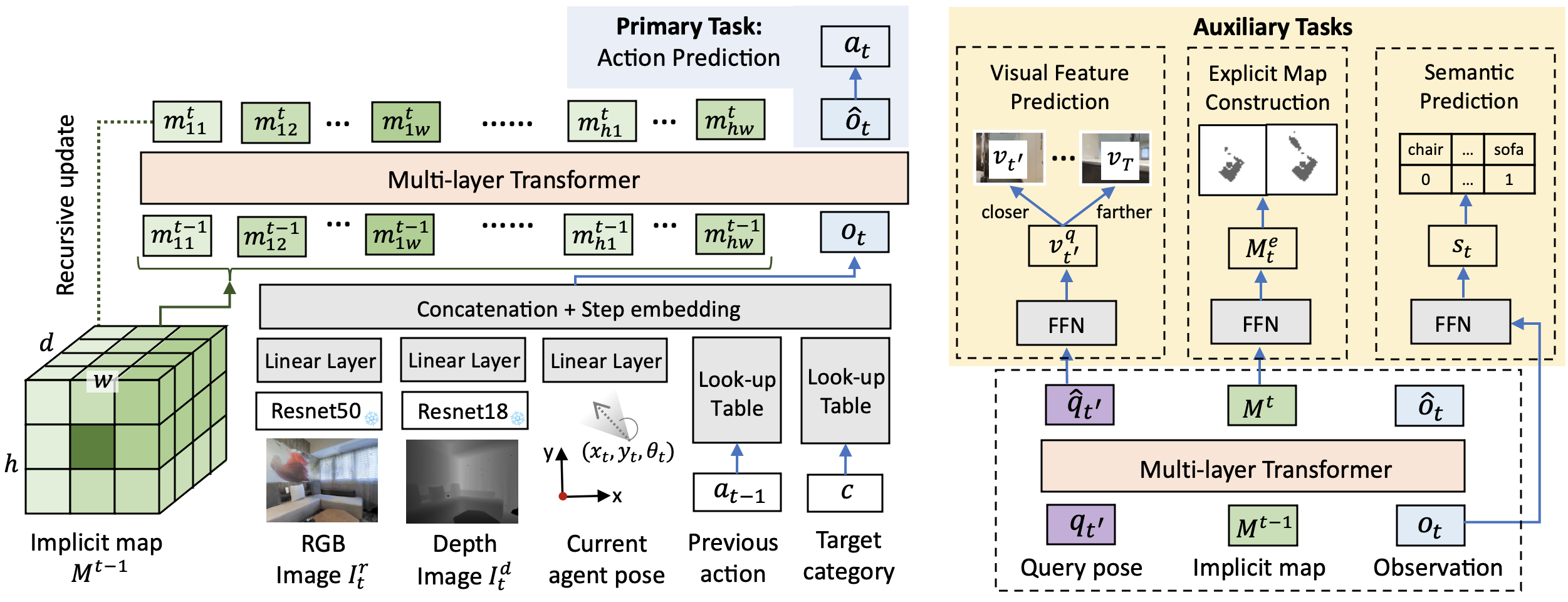}
    \caption{An overview of our ObjectNav model with a recursive implicit map (RIM). It encodes observations at step $t$ into $o_t$, and uses a multi-layer transformer to recursively update the implicit spatial map $M^t$. Besides the main task of action prediction, three auxiliary tasks are proposed to improve spatial reasoning of $M^t$ and semantic understanding. The Multi-layer Transformer on the right is the same as on the left, while we add three feed-forward networks (FFN) for auxiliary tasks.
    }
    \label{fig:method}
    \vspace{-1.5em}
\end{figure*}

\subsection{Task Definition}
\label{sec:method_task}

The object goal navigation task~\cite{batra2020objectnav} requires an agent to navigate to an instance of a given object category in unseen environments, such as ``chair'' or ``table''.
The agent is initialized at a random navigable location and asked to find the goal object category. 
It is equipped with an RGB-D camera capturing an RGB image $I^r_t$ and a depth image $I^d_t$ at each step $t$. 
It also receives the agent's pose relative to its start position including $(x_t, y_t)$ coordinates and orientation $\theta_t$ in a 2D plane. 
The agent should execute an action $a_t \in \mathcal{A}$ at each step, where $\mathcal{A}$ is an action space consisting of \verb|move_forward| for 0.25 meters, \verb|turn_left|, \verb|turn_right|, \verb|look_up|, \verb|look_down| with rotation angle of 30 degrees, and \verb|stop|.
The episode terminates if the agent executes the \verb|stop| action or if it runs out of a budget of $T=500$ steps.
It is considered successful when the agent predicts \verb|stop| at a location where the distance to the target object is less than 1 meter and the target object can be viewed. %

\subsection{Navigation Model with Recursive Implicit Map}
\label{sec:method_model}

We propose an implicit map representation to encode episodic navigation history into a compact spatial memory. This implicit map is recursively updated and used to predict actions for object goal navigation.
Figure~\ref{fig:method} (left) illustrates an overview of the model and its three modules: observation encoding, implicit map updating and action prediction.

\noindent\textbf{Observation Encoding.}
We encode the observation into a feature vector $o_t$ at each step, including RGB image $I^r_t$, depth image $I^d_t$, agent's pose $(x_t, y_t, \theta_t)$, previous action $a_{t-1}$ and  target object category $c$.
To be specific, we use a pretrained CLIP Resnet50 \cite{radford2021learning} to encode $I^r_t$ and Resnet50 pretrained in PointNav~\cite{wijmans2019dd} to encode $I^d_t$. 
The visual encoders stay frozen as in~\cite{khandelwal2022simple} to make the training more efficient and enable generalization to unseen environments. %
The agent's pose is converted to a feature $(x_t, y_t, sin \theta_t, cos \theta_t)$.
The visual and pose features are then passed through their own linear layer.
The $a_{t-1}$ and $c$ are embedded through two learnable look-up tables~\cite{ramrakhya2022habitatweb} respectively.
We concatenate all the features and add a sinusoidal positional embedding~\cite{vaswani2017attention} of the current step $t$ to obtain the final observation feature $o_t$.

\noindent\textbf{Implicit Map Representation and Updating.}
We represent the navigation history until step $t$ as an implicit spatial map $M^t = [m^t_{ij}]_{h \times w}$ with $h \times w$ cells.
Each cell $(i, j)$ consists of a latent feature $m^t_{ij} \in \mathbb{R}^{d}$ where $d$ is the feature dimensionality and its position is computed relatively to the center of the map as $(i-\frac{h}{2}, j-\frac{w}{2})$. 
At the beginning of an episode, we initialize $M^0$ using the cell position: 
\begin{equation}
\label{eqn:imap_init}
    m^0_{ij} = w^{0}_{m} + \text{FFN}([i-\frac{h}{2}, j-\frac{w}{2}]),
\end{equation}
where $w^0_m \in \mathbb{R}^d$ is a learnable embedding and $\text{FFN}$ is a feed-forward network.
Then, at each step $t$, the implicit map is updated given the new observation $o_t$ with a differentiable function. Here we employ a multi-layer transformer~\cite{vaswani2017attention} due to its effectiveness in many computer vision and robotics tasks~\cite{dosovitskiy2020image,chen2021history,guhur2022instruction,brohan2022rt}.
We concatenate all the map cells and the encoded observation as tokens to feed in the transformer.
The core of the transformer layer is the self-attention block to compute relations among different tokens:
\begin{equation}
    \text{SelfAttn}(X) = \text{Softmax}\left(\frac{XW_q (XW_k)^{T}}{\sqrt{d}}\right) XW_v,
    \label{eqn:attn}
\end{equation}
where $X=[m^{t-1}_{11}, \cdots, m^{t-1}_{hw}, o_t]$. 
To enhance the geometry alignment between our implicit map and the observation, we explicitly add the corresponding position feature to the map and observation token embeddings inside the Softmax function of self-attention. 
We denote the output embeddings as $M^{t}$ and $\hat{o}_t$ for the map tokens and observation token respectively.
The $M^t$ is used as the input for the next step.
Compared to explicit map representations \cite{chaplot2020object,chaplot2020neural,jatavallabhula2023conceptfusion}, the construction of our implicit map does not require precise localization and sensors since it learns how to update the map based on visual and geometric features.
Compared to existing implicit representations \cite{ramrakhya2022habitatweb,pashevich2021episodic}, our implicit map is geometry-aware and more compact.

\noindent\textbf{Action Prediction.}
During the update of the implicit map, the observation token $o_t$ is also updated through the transformer layer into $\hat{o}_t$.
We then directly employ $\hat{o}_t$ to predict the low-level action $p(a_{t}) = \text{Softmax}(\text{FFN}(\hat{o}_t))$ where FFN is a single-layer feed-forward network.

\subsection{Training Objectives}
\label{sec:method_train}

\noindent\textbf{Primary task.}
We utilize behavior cloning to train the navigation model.
Given an expert trajectory $\tau=(o_0,a^{*}_0, \cdots, o_T,a^{*}_T)$, we use the cross-entropy loss with inflection weighting~\cite{wijmans2019embodied} for action prediction, which gives higher weights for actions different from the previous one. The loss for the trajectory is as follows:
\begin{equation}
\label{eqn:action_prediction}
    L_{\text{AP}} = \frac{1}{T} \sum\nolimits_{t=0}^{T} - (1 + \gamma \mathds{1}_{a_t^* \neq a_{t-1}^*}) \text{log}~p(a^{*}_t).
\end{equation}

\noindent\textbf{Auxiliary tasks.}
We propose three auxiliary tasks to further improve spatial reasoning of the implicit map and object recognition (Figure~\ref{fig:method} right): visual feature prediction, explicit map construction and semantic prediction.

The visual feature prediction task aims to predict a visual feature of the agent's observation given the agent's pose and the implicit map, which can enhance memorizing the scene. 
Specifically, we randomly select a pose $(x_{t'}, y_{t'}, \theta_{t'})$ in the trajectory $\tau$ as the query pose, where $t' \in [0, t+k]$ at each step $t$. We empirically use $k=20$ to encourage the ability to imagine the near future.
Suppose $q_{t'}$ is the pose of the agent encoded with a linear layer, and $v_{t'}$ is the visual semantic feature at step $t'$ encoded with CLIP~\cite{radford2021learning}. 
We concatenate $q_{t'}$ with the map and observation tokens and feed them to the transformer as illustrated in the bottom right of Figure~\ref{fig:method}.
As we only want $q_{t'}$ to extract latent features from the implicit map but not to affect the update of the map, we use an attention masking trick to prevent the attention from $M^t$ and $o^t$ to $q_{t'}$. 
The output embedding $\hat{q}_{t'}$ is fed into a feed-forward network to predict a visual feature as $v^q_{t'}$.
We employ the contrastive loss to make $v^q_{t'}$ closer to the original CLIP feature $v_{t'}$ at step $t'$ while increasing the distance between $v_{t'}$  and features at other locations on the trajectory:
\begin{equation}
\label{eqn:visual_prediction}
    L_{VP} = \frac{1}{T} \sum\nolimits_{t=0}^{T} \frac{s(v^q_{t'}, v_{t'})}{\sum_{i=0}^{t+k} s(v^q_{t'}, v_i)},
\end{equation}
where $s(\cdot, \cdot)$ is the cosine similarity.

Our second auxiliary task
aims to derive 2D occupancy maps given the implicit representation, which can help encode the geometry of the scene. We generate groundtruth explicit maps $M^t_e \in \mathbb{R}^{H \times W \times 2}$ using the depth images and camera poses~\cite{chaplot2020object}, where we use $H = W = 48$ with each cell corresponding to a 25$\times$25cm area. The first channel denotes whether the cell is occupied by obstacles while the second channel denotes whether the cell has been viewed. The agent is at the center of $M^t_e$ at each step $t$. 
Then, we employ a feed-forward network (FFN) to predict $M^t_e$ given $M^t$ with the following loss function:
\begin{equation}
    L_{EM} = \frac{1}{T} \sum\nolimits_{t=0}^{T} \text{BCE}(\text{FFN}(M^t), M^t_e),
\end{equation}
where $\text{BCE}$ is the binary cross entropy loss.

In the third auxiliary task of semantic prediction, we use the encoded RGB-D features to predict the existence and object mask to image size ratio for each object category at each step. The groundtruth labels are available in the training environments. %
We compute the binary cross entropy loss for semantic prediction denoted as $L_{SP}$.

The overall training objective is a weighted combination of losses for the main task and the three auxiliary tasks:
\begin{equation}
\label{eqn:overall_loss}
    L = L_{AP} + \lambda (L_{VP} + L_{EM} + L_{SP}).
\end{equation}

\section{Experiments}
\label{sec:expr}

\subsection{Experimental Setting}

\noindent\textbf{Datasets.}
We use human demonstrations collected by Habitat-web~\cite{ramrakhya2022habitatweb} in  Matterport3D (MP3D) environments \cite{chang2017matterport3d} for training, which contain 70k trajectories in 56 training environments in MP3D.
These demonstrations are obtained by manually controlling a virtual robot in the Habitat simulator~\cite{savva2019habitat} given egocentric images with horizontal field of view (HFoV) of 79 degrees and vertical field of view (VFoV) of 63 degrees. The virtual robot has a radius of 18cm and height of 88cm.
The target object categories in the demonstrations include 21 real-world objects
\footnote{The full list of the real-world objects is ``chair'', ``sofa'', ``plant', ``bed'', ``toilet'', ``tv monitor'', ``table'', ``sink'', ``picture'', ``shower'', ``bathtub'', ``stool'', ``cabinet'', ``cushion'', ``chest of drawers'', ``bathtub'', ``counter'', ``fireplace'', ``gym equipment'', ``seating'', ``towel'' and ``clothes''.} 
and 6 synthetic objects.
To be noted, the RGB-D images are imperfect in simulation due to incomplete 3D reconstruction of the environments whereas the poses of agents are exact.
We report results on the MP3D validation split, which consists of 2,195 episodes from 11 unseen environments for the 21 real-world object categories. 
To further measure the generalization of our model, we evaluate the model trained on MP3D on the HM3D~\cite{yadav2022habitat} dataset, which contains more diverse and cluttered navigation scenes than MP3D but only focuses on \csz{6 dominant object categories~\footnote{The list of objects is ``chair'', ``couch'', ``plant'', ``bed'', ``toilet'' and ``tv''.} in the 21 categories of MP3D.
} 

\noindent\textbf{Evaluation metrics.}
We use three main metrics~\cite{batra2020objectnav} to evaluate the object goal navigation performance.  SR (Success Rate) is the ratio of episodes where the agent successfully reaches the target object.
SPL (Success rate weighted by Path Length) measures navigation efficiency by dividing SR by the path length relative to the shortest path length.
SoftSPL is a softer version of SPL which reports the progress towards the goal (the ratio of reduced distance to the initial distance) and penalizes it by the path length.

\noindent\textbf{Implementation details.}
We use linear layers with hidden sizes of 128, 256, 64, 32, 32 to encode RGB, depth, agent's pose, previous action and target category features respectively.
A 4-layer transformer with hidden size of 512 is used to update the implicit map.
We train models for 25 epochs with a batch size of 32. The learning rate is initialized as 0.0003 with linear decrease. 
We use the AdamW optimizer with weight decay 0.01. The inflection weight $\gamma$ in Eq~(\ref{eqn:action_prediction}) is set to 3.48 as in previous work~\cite{ramrakhya2022habitatweb}, and the loss balance factor $\lambda$ in Eq~(\ref{eqn:overall_loss}) is set to 0.3.
At inference time, we select the action with the highest probability at each step. 
We also use the agent's pose to detect collisions in the simulation. If the agent collides, we select the second best action at that step.

\subsection{Ablation Studies}

We perform extensive experiments on the MP3D validation split to demonstrate the effectiveness of the proposed recursive implicit map for object goal navigation.

\noindent\textbf{Comparison with existing implicit representations.}
We compare with two state-of-the-art implicit representations:
a recurrent state learned in recurrent neural networks~\cite{maksymets2021thda,Ye_2021_ICCV,ramrakhya2022habitatweb,khandelwal2022simple} and an episodic sequence encoded by transformers~\cite{pashevich2021episodic,chen2021history}.
For fair comparison, we use the same observation encoding and the objective in Eq~(\ref{eqn:action_prediction}) to train all these models.
Table~\ref{tab:memory_cmpr} presents the results.
The recurrent state is insufficient to encode the navigation history and achieves the poorest performance under all metrics.
The episodic sequence memory significantly improves the object search performance with 5.56 absolute gain on SR, which is in line with previous observations for visual navigation~\cite{pashevich2021episodic,chen2021history}. However, it does not capture the geometry structure of the environment and its computation cost continuously grows with the number of steps.
Our recursive implicit map, instead, learns a geometry-aware and more compact representation of the explored environment. It outperforms episodic sequence by 6.7\% relative gain on SR and SPL, and runs 1.36$\times$ faster at inference on the MP3D dataset.

\begin{table}
\centering
\caption{Performance of different implicit representations on the MP3D val split, where $d$ is the feature dimensionality, $T$ is the episode length, and $h, w$ is the size of our implicit map. Here we use $d=512, T=500, h=w=3$.}
\label{tab:memory_cmpr}
\begin{tabular}{lcccc} \toprule
 & Memory size & SR & SPL & SoftSPL \\ \midrule
Recurrent state & $1 \times d$ & 38.95 & 11.09 & 16.35 \\
Episodic sequence & $T \times d$& 44.51 & 14.17 & 19.35 \\
Recursive implicit map & $h \times w \times d$ & \textbf{47.74} & \textbf{15.12} & \textbf{20.51}  \\ \bottomrule
\end{tabular}
\vspace{-1em}
\end{table}

\noindent\textbf{Size of the implicit map.}
In Table~\ref{tab:imap_size}, we evaluate the influence of the map size on the navigation performance.
We can see that our model with the smallest $1 \times 1$ map already outperforms the recurrent state that has the same memory size and the episodic sequence with a much larger memory size in Table~\ref{tab:memory_cmpr}.
The result demonstrates that the recursive transformer architecture is more effective than recurrent neural networks to encode long range sequences. It also indicates that there exists large redundancy in the episodic sequence due to the small navigation steps and it is unnecessary to keep all the steps in the memory.
Increasing the size of our recursive implicit map from 1 to 3 and 5 continuously improves the navigation performance. However, the performance saturates with a larger size of 7.
As the initial distance to goal in MP3D is about 8 meters on average, it is reasonable that a small map size is capable of modeling the explored environment. 
Since the map of size $3 \times 3$ already outperforms the other implicit representations and is more efficient than larger sizes, we use $3 \times 3$ size by default in the following experiments.

\begin{table}
\centering
\caption{Performance of recursive implicit map with different sizes and position encodings on the MP3D val split.}
\label{tab:imap_size}
\begin{tabular}{cccccc} \toprule
Map size & Map pos & Agent pos & SR & SPL & SoftSPL \\ \midrule
 $1 \times 1$ & \checkmark & \checkmark& 45.74 & 14.94 & 19.54 \\
$3 \times 3$ & \checkmark & \checkmark & 47.74 & 15.12 & 20.51 \\
$5 \times 5$ & \checkmark & \checkmark & \textbf{49.52} & 15.84 & \textbf{21.43} \\
$7 \times 7$ & \checkmark  & \checkmark & 48.52 & \textbf{15.86} & 21.23 \\ \midrule
\multirow{2}{*}{$3 \times 3$} & $\times$  & \checkmark & 46.70 & 15.15 & 20.46 \\ 
 & \checkmark & $\times$ & 47.43 & 13.90 & 18.77 \\ 
\bottomrule
\end{tabular}
\vspace{-2em}
\end{table}

\noindent\textbf{Position encoding of the map and agent.}
The position of the agent and the observed environment plays a critical role in explicit map construction. Hence, we evaluate the impact of position of the map and the agent on our model with an implicit map.
In the bottom block of Table~\ref{tab:imap_size}, we either replace the explicit position of each map cell by learnable embeddings or remove the agent's pose features in the model.
We can see that both the map position and agent position are beneficial, but the model without the position information especially the agent's pose still performs well as it can implicitly update the spatial memory via visual features. The robustness to agent's pose is an important property since it is usually difficult to get accurate robot pose in the real world.

\begin{figure}
     \centering
     \includegraphics[width=\linewidth]{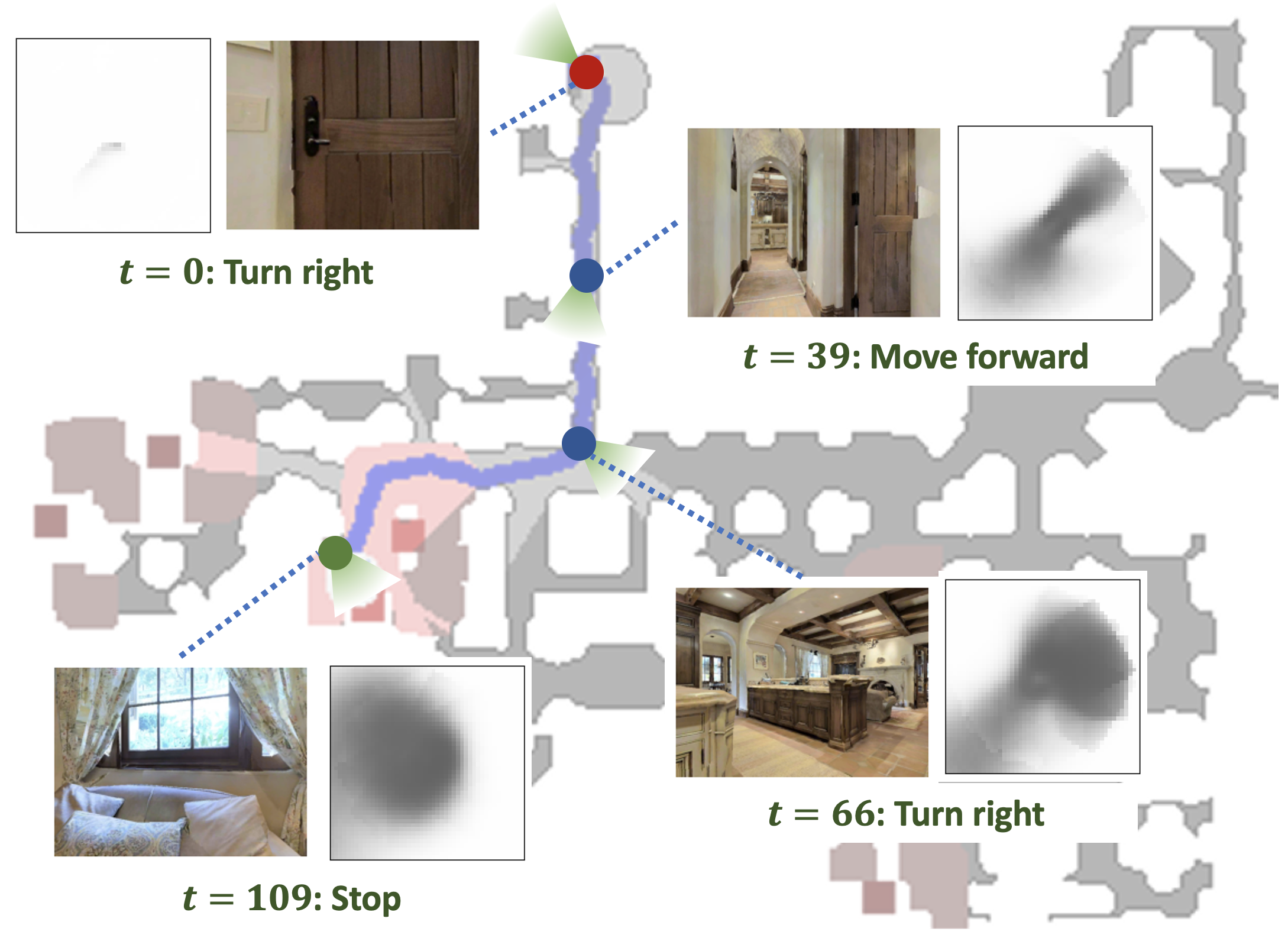}
     \caption{Predicted navigation trajectory (the blue curve) for finding a target object `\textbf{sofa}' on the MP3D val split. The red and green circles denote the start and stop positions respectively. Blue circles are points along the way where $t$ denotes the discrete navigation step.
     We visualize the egocentric RGB image and the predicted exploration map from our implicit representation at each selected point.}
    \label{fig:examples}
    \vspace{-1em}
\end{figure}

\noindent\textbf{Auxiliary tasks in training.}
Table~\ref{tab:aux_tasks} presents an ablation analysis for the three proposed auxiliary tasks.
The first row is the $3 \times 3$ model in the upper block of Table~\ref{tab:imap_size}.
The visual feature prediction task contributes most to navigation efficiency with about 11\% relative gain on SPL and SoftSPL, as it strengthens the implicit map to infer visual representations of the scene.
The explicit mapping task has a slight but stable improvement on all  three metrics.
The semantic prediction task improves the success rate in particular due to better object recognition.
The three auxiliary tasks are also complementary to each other, and their combinations achieve an additional performance boost as shown in the last row.
Figure~\ref{fig:examples} visualizes the navigation result of our best model for finding a target object `sofa' on the MP3D val split. 

\begin{table}
\centering
\caption{Performance of training with different auxiliary tasks on the MP3D val split.}
\label{tab:aux_tasks}
\begin{tabular}{cccccc} \toprule
\begin{tabular}[c]{@{}c@{}}Visual feature \\ prediction\end{tabular} & \begin{tabular}[c]{@{}c@{}}Explicit \\ mapping\end{tabular} & \begin{tabular}[c]{@{}c@{}}Semantic \\ prediction\end{tabular} & SR & SPL & SoftSPL \\ \midrule
$\times$ & $\times$ & $\times$ & 47.74 & 15.12 & 20.51 \\
\checkmark & $\times$ & $\times$ & 47.84 & 16.86 & 22.71 \\
$\times$ & \checkmark & $\times$ & 48.06 & 15.81 & 20.86 \\
$\times$ & $\times$ & \checkmark & 48.66 & 15.59 & 20.10 \\
\checkmark & $\times$ & \checkmark & 48.97 & \textbf{17.86} & \textbf{23.07} \\
\checkmark & \checkmark & \checkmark & \textbf{50.25} & 17.00 & 22.62 \\ \bottomrule
\end{tabular}
\vspace{-1em}
\end{table}

\subsection{Comparison with the State of the Art}

\noindent\textbf{MP3D dataset.}
Table~\ref{tab:mp3d_sota} presents the comparison with state-of-the-art methods on the MP3D dataset.
We compare with both explicit mapping methods and implicit representation based methods.
The explicit mapping methods build a 2D map for exploration and are combined with a pretrained object segmentation model for perception. Once the target object is detected, the agent navigates to the target using the map or a PointNav policy.
FBE~\cite{yamauchi1997frontier} is a classical frontier-based exploration approach which builds an occupancy map and navigates to the nearest map frontiers; ANS~\cite{chaplotlearning} is a neural SLAM based exploration policy trained by RL to maximize area coverage;
SemExp~\cite{chaplot2020object} builds a semantic map to predict the next goal trained by RL;
PONI~\cite{ramakrishnan2022poni} is similar to SemExp~\cite{chaplot2020object} but uses behavior cloning for training; and STUBBORN~\cite{luo2022stubborn} improves SemExp~\cite{chaplot2020object} by accumulating semantic segmentation results across steps for object recognition and utilizing a multi-scale occupancy map for path planning.
Different from explicit mapping approaches which require hand-crafted map design and engineering, the implicit representation based methods \cite{maksymets2021thda,Ye_2021_ICCV,khandelwal2022simple,ramrakhya2022habitatweb} train end-to-end models to predict actions, in particular the recurrent neural network.
Our model achieves significant improvement over the existing explicit mapping and implicit representation based methods.
We also submitted our model (RIM) to the hidden test split of MP3D and achieved the first place on the leaderboard\footnote{\url{https://eval.ai/web/challenges/challenge-page/802/leaderboard/2195}, 1st March, 2023.}.
Note that there is a large performance gap between the validation and test splits, which we suspect might result from different category distributions as many objects are long-tailed.

\begin{table}
\centering
\caption{Comparison with state-of-the-art methods on the MP3D val and test-standard split.}
\label{tab:mp3d_sota}
\begin{tabular}{llcccc} \toprule
\multirow{2}{*}{} & \multirow{2}{*}{Method} & \multicolumn{2}{c}{Val} & \multicolumn{2}{c}{Test-standard} \\
&  & SR & \multicolumn{1}{c}{SPL} & SR & \multicolumn{1}{c}{SPL} \\ \midrule
\multirow{5}{*}{\begin{tabular}[c]{@{}c@{}}Explicit\\ mapping\end{tabular}} & FBE~\cite{yamauchi1997frontier} & 20.0 & 7.6 & - & - \\
& ANS~\cite{chaplotlearning} & 21.2 & 9.4 & - & -\\
& SemExp~\cite{chaplot2020object} & - & - & 17.9 & 7.1 \\
& PONI \cite{ramakrishnan2022poni} & 27.8 & 12.0 & 20.0 & 8.8 \\ 
& STUBBORN~\cite{luo2022stubborn} & 37.0 & - & 23.7 & 9.8 \\ \midrule
\multirow{6}{*}{\begin{tabular}[c]{@{}c@{}}Implicit\\ representation\end{tabular}} & THDA \cite{maksymets2021thda} & 28.4 & 11.0 & 21.4 & 8.9 \\
& Red Rabbit \cite{Ye_2021_ICCV} & 34.6 & 7.9 & 23.7 & 6.2 \\
& EmbCLIP \cite{khandelwal2022simple} & 21.6 & 8.7 & 18.1 & 7.8 \\
& OVRL~\cite{yadav2022offline} & 28.6 & 7.4 & 24.9 & 8.3 \\
& Habitat-web \cite{ramrakhya2022habitatweb} & 35.4 & 10.2 & 27.8 & 9.9 \\ \cmidrule{2-6}
& RIM (Ours) & \textbf{50.3}  & \textbf{17.0} & \textbf{37.6} & \textbf{15.6} \\ \bottomrule
\end{tabular}
\vspace{-1.5em}
\end{table}

\noindent\textbf{Generalization to HM3D dataset.}
To further measure the generalization ability of our model, we directly evaluate our model trained on MP3D human demonstrations on the HM3D dataset.
Table~\ref{tab:hm3d} presents the cross-dataset performance.
Our model significantly outperforms other methods in cross-dataset generalization as shown in the bottom block.
Although our model is not fine-tuned on the HM3D dataset which has more diverse buildings and focuses on a subset of goal categories in MP3D, RIM still achieves competitive performance when compared to other methods trained on HM3D.
We submitted our model to the test-standard  split\footnote{\url{https://eval.ai/web/challenges/challenge-page/1615/leaderboard/3899}, 1st March, 2023.} of the HM3D dataset.
The gap between val and test splits is smaller compared to the MP3D dataset, as the HM3D dataset focuses on 6 dominant object categories.

\begin{table}
\centering
\tabcolsep=0.15cm
\caption{Comparison with state-of-the-art methods on the HM3D val and test-standard split. Top rows: in-domain training on the HM3D dataset. Bottom rows: cross-domain generalization with training on the MP3D dataset. 
}
\label{tab:hm3d}
\begin{tabular}{ccccccc} \toprule
\multirow{2}{*}{\begin{tabular}[c]{@{}c@{}}Training\\ scenes\end{tabular}} & \multirow{2}{*}{Method} & \multirow{2}{*}{\begin{tabular}[c]{@{}c@{}}Training\\ algorithm\end{tabular}}  & \multicolumn{2}{c}{Val} & \multicolumn{2}{c}{Test-standard} \\
&  & & SR & \multicolumn{1}{c}{SPL} & SR & \multicolumn{1}{c}{SPL} \\ \midrule
\multirow{5}{*}{\begin{tabular}[c]{@{}c@{}}HM3D\end{tabular}} & DD-PPO~\cite{habitatchallenge2022} & RL & 27.9 & 14.2  & 26.0 & 12.0 \\
 & Habitat-web~\cite{ramrakhya2022habitatweb} & BC & 57.6 & 23.8 & 55.0 & 22.0 \\
 & ProcTHOR~\cite{deitke️2022procthor} & RL & - & - & 54.0 & 32.0 \\
 & OVRL~\cite{yadav2022offline} & RL & \textbf{62.0} & 26.8 & 60.0 & 27.0 \\ 
& \textcolor{gray}{ByteBOT}\footnotemark & - & - & - & \textcolor{gray}{\textbf{68.0}} & \textcolor{gray}{\textbf{37.0}} \\  

\midrule

\multirow{4}{*}{\begin{tabular}[c]{@{}c@{}}MP3D\end{tabular}} & DD-PPO~\cite{habitatchallenge2022} & RL & 19.0 & 6.9 & - & - \\
& Habitat-web~\cite{ramrakhya2022habitatweb} & BC & 25.4 & 7.7 & - & - \\
& SemExp~\cite{chaplot2020semantic} & RL & 50.8 & 21.7 & - & - \\ \cmidrule{2-7}
& RIM (Ours) & BC & 57.8 & \textbf{27.2} & 61.4 & 29.3 \\ \bottomrule
\end{tabular}
\vspace{-1.5em}
\end{table}
\footnotetext{Unpublished work submitted to the HM3D test-standard leaderboard.}

\subsection{Real Robot Experiments}

Besides demonstrating the state-of-the-art performance in simulation, we deploy our model on a real robot. 

\begin{figure}
    \centering
    \includegraphics[width=1\linewidth]{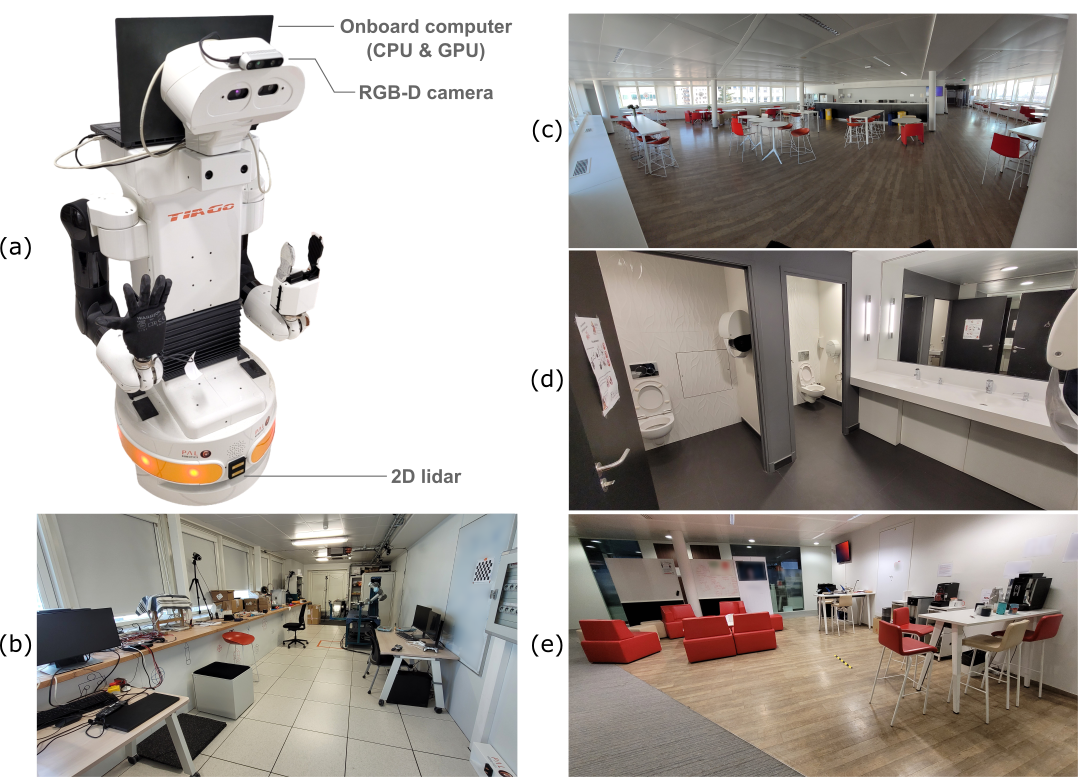}
    \caption{\textbf{Robotic setup.} We perform real-world experiments with a Tiago++ robot (a) in various environments: a robotics room (b), a large canteen (c), toilets (d) and a lounge (e).
    }
    \label{fig:real_world}
    \vspace{-1em}
\end{figure}

\noindent\textbf{Robotic setup.}
We use a Tiago++\footnote{\url{https://pal-robotics.com/robots/tiago/}} mobile robot as shown in Figure~\ref{fig:real_world}a. We equip the robot with an onboard computer (including an Intel i9-10980HK CPU and a GeForce RTX 2080 GPU) and an Intel RealSense D435 RGB-D camera, whose horizontal field of view (HFoV) is about 55 degrees and vertical field of view (VFoV) is 42 degrees. 
The robot has a radius of 27cm and a minimum height of 110cm. We set its linear and angular velocities to 0.2m/s and 0.2rad/s respectively. To prevent collisions with the environment, we exploit a 2D lidar placed at 10cm above the ground: when the laser range gets under 50cm from the robot and the predicted action is moving forward, we select the next best action and continue the episode. However, if during an episode the robot collides with unmovable obstacles, which can happen with objects that go undetected for a 2D lidar (e.g. tables), we force the robot to stop and treat such episodes as failure.
This makes the real-world evaluation more challenging than that in the simulator where current benchmarks do no punish collisions. 
Our testing environment is an office building with one robotics room, one large canteen, toilets and one lounge (see Figure~\ref{fig:real_world}b-e). We evaluate real-robot ObjectNav for 7 object categories including ``chair'', ``cushion'', ``plant'', ``sink'', ``sofa'', ``table'' and  ``toilet''.

\noindent\textbf{Model retraining in simulation.}
The training demonstrations in simulation from~\cite{ramrakhya2022habitatweb} are collected using a robot with radius of 18cm and a camera with height of 88cm, HFoV of 79$^\circ$ and VFoV of 63$^\circ$. Since we have different robot and camera configurations, we need to re-generate demonstrations in the simulator for policy learning.
However, enlarging the robot height and radius to replicate our Tiago++ robot would make the original human demonstrations invalid due to collisions \emph{e.g.} bumping into high obstacles or being unable to pass narrow passages.
Therefore, we mount a virtual camera on the original simulated robot with a height of 120cm and HFoV and VFoV of 55 and 42 degrees respectively, keeping the robot size unchanged. We use the RGB-D images from the new camera in training.
In addition, as obtaining accurate agent's poses is hard in the real world, we simply discard the agent's pose input to our model, as our model is robust even without the agent's pose as shown in Table~\ref{tab:imap_size}.
Table~\ref{tab:sim_camera_change} presents results for the MP3D val split using the new camera parameters. With such changes, the performance significantly decreases and we observe there are much more collisions due to limited VFoV (second row).
To make the view of the ground with the new camera configuration similar to the initial one used in the simulation experiments, we tilt the camera by 20 degrees below horizon. We can see from the last row in Table~\ref{tab:sim_camera_change} that the tilt is very beneficial, leading to even better performance than the original robotic setup in MP3D evaluation setup.

\begin{table}
\centering
\caption{Performance on MP3D val split with different camera parameters. 
The pose of the agent is discarded here.}
\vspace{-0.2em}
\label{tab:sim_camera_change}
\begin{tabular}{cccccc} \toprule
\begin{tabular}[c]{@{}c@{}}Height \\ (cm)\end{tabular} & \begin{tabular}[c]{@{}c@{}}HFoV $\times$ VFoV \\ (deg $\times$ deg)\end{tabular} & \begin{tabular}[c]{@{}c@{}}Tilt \\ (deg)\end{tabular} & SR & SPL & SoftSPL \\ \midrule
88 & 79 $\times$ 63 & 0 & 47.43 & 13.90 & 18.77 \\
120 & 55 $\times$ 42 & 0 & 40.36 & 12.48 & 19.36 \\
120 & 55 $\times$ 42 & -20 & 47.43 & 16.65 & 21.68 \\ \bottomrule
\end{tabular}
\vspace{-2.5em}
\end{table}

\noindent\textbf{Model fine-tuning in real environments.}
Our testing environments are quite different from the training scenes, both in terms of the space size and object appearance.
To bridge this sim-to-real gap, we manually collected 70 demonstrations on the actual robot with 10 demonstrations per object.
\csz{We collect 5 episodes for chairs, cushions, plants and tables respectively in the robotics room; 5 episodes for chairs, cushions, plants, tables and sinks in the canteen; 5 for sinks and 10 for toilets in two restrooms; and 10 sofas in the lounge.}
We fine-tune the navigation policy trained in simulation with these collected real-world demos via behavior cloning for 120 iterations with a batch size of 32 episodes. 
All parameters except those in the RGB-D image encoders are updated.
We additionally obtain two model variants by fine-tuning on two subsets. 
\csz{The first one uses 20 demos collected only in the robotics room, including 5 demos for 4 objects, 
i.e. chair, cushion, plant and table.} 
This allows us to  evaluate generalization to new environments and objects in the real world. 
\csz{The second subset uses half of the 70 demos by randomly selecting 5 episodes from the 10 episodes per object}, to measure improvements due to more in-domain demonstrations.

\noindent\textbf{Results in the real world.}
We evaluate the fine-tuned policies on the Tiago++ robot in the real world. 
\csz{For each object category, we run 5 episodes in each room where the object is located, e.g. 5 in the robotics room and 5 in the canteen for chairs, except for sofas that are only found in the lounge and for which we run 10 episodes in that same location. For each episode, we randomly select the initial positions of objects and the robot in the environment.}
In our experiments, the initial distance to the nearest target object is 3.2$\pm$1.4m and in 88.6\% of the episodes the robot cannot see any goal object from its initial position, thus requiring exploration.
Our model is efficient: it takes 46$\pm$5ms to predict the action at each step and uses 1.9Gb of GPU memory. 

\begin{table}
\centering
\tabcolsep=0.15cm
\vspace{0.2em}
\caption{Performance in the real world (\# of success / \# of testing episodes). We fine-tune policies with different numbers of real-world demonstrations, and evaluate with and w/o an additional pretrained semantic segmentation model. 
}
\label{tab:real_robot_expr}
\vspace{-0.2em}
\begin{tabular}{cccccccccc} \toprule
Data & Segm & Chair & Cushion & Plant & Table & Sink & Sofa & Toilet\\ \midrule
\multirow{2}{*}{20} & $\times$   & 9/10 & 2/10 & 3/10 & 8/10 & 3/10 & 2/10 & 3/10\\
                    & \checkmark & 9/10 & 4/10 & 4/10 & 7/10 & 6/10 & 4/10 & 3/10 \\ \midrule
 \multirow{2}{*}{35} & $\times$   & 10/10 & 3/10 & 5/10 & 7/10 & 6/10 & 8/10 & 4/10\\
                     & \checkmark & 10/10 & 3/10 & 6/10 & 5/10 & 6/10 & 8/10 & 6/10 \\ \midrule
\multirow{2}{*}{70} & $\times$   & 8/10 & 1/10 & 2/10 & 8/10 & 2/10 & 8/10 & 5/10\\
                    & \checkmark & 8/10 & 1/10 & 2/10 & 5/10 & 5/10 & 9/10 & 2/10 \\ \bottomrule
\end{tabular}
\vspace{-2.5em}
\end{table}

Table~\ref{tab:real_robot_expr} shows the results. 
The model fine-tuned with only 20 demonstrations can achieve some generalization to new rooms and new objects, i.e., sink, sofa and toilet, but its accuracy is somewhat limited. 
\csz{Increasing the number and diversity of demonstrations (the second block in Table~\ref{tab:real_robot_expr}) improves the success rate for most categories, in particular the ones not seen by the previous model. %
In addition, we find that the policy trained only in the robotics room tends to explore less than those trained in all the environments. This is because the robotics room is small, leading to short and less exploratory trajectories during our data collection.
However, the model trained with all 70 demonstrations underperforms the model trained with only half of the data. 
This suggests that fine-tuning too much may reduce the generalization ability of pretrained policies in simulation.
}

To reduce the error of failing to stop, we further combine our navigation policy with a state-of-the-art object segmentation model~\cite{chen2022vitadapter} pretrained on the ADE20K real image dataset: we multiply the segmentation mask of the detected target object with the depth image to obtain the distance to the target object, and stop the episode if the median distance is below 1m and the object area ratio in the image is above 1\%. 
We observe that this addition often leads to improved results, see rows with segmentation  in Table~\ref{tab:real_robot_expr}.
\csz{
But it also has the following problems: i) it cannot help if the agent does not get close to a target object; ii) it suffers from noisy depth images when computing the distance to an object; iii) its predictions are less accurate due to a limited view of the scene, e.g., treating counters as tables; and iv) it significantly affects the inference speed and GPU memory.
}

Our real world experiments demonstrate several limitations of the approach.  First, as the robot size is larger than that the one used in simulation, the policy is prone to collide on edges. 
Second, the discrete action space is too restrictive, making it hard for the robot to go through narrow passages like restrooms with doors to find a toilet. 
Third, we observe a large sim-to-real gap for the visual observations. The MP3D dataset contains incomplete 3D meshes, generating observations with holes which may impact real-world performances.

\section{Conclusion}
\label{sec:conclusion}

This paper proposes an implicit spatial map to improve end-to-end learning methods for object goal navigation.
The implicit map is recursively updated with new observations  using a transformer.
We propose three auxiliary tasks of visual feature prediction, explicit map construction and semantic prediction to improve the learning of the implicit map.
Our method outperforms the state of the art in the challenging MP3D dataset, and generalizes well to the HM3D dataset.
We also conduct real robot experiments and achieve promising results by fine-tuning with a few real-world examples.

{
\small
\textbf{Acknowledgement}:
This work was granted access to the HPC resources of IDRIS under the allocation 101002 made by GENCI. 
It was funded in part by the French government under management of Agence Nationale de la Recherche as part of the “Investissements d’avenir” program, reference ANR19-P3IA-0001 (PRAIRIE 3IA Institute), the ANR project VideoPredict (ANR-21-FAI1-0002-01) and by Louis Vuitton ENS Chair on Artificial Intelligence.
}

\bibliographystyle{IEEEtran}

\end{document}